\pdfoutput=1

\documentclass[11pt]{article}

\usepackage[preprint]{acl}

\usepackage{times}
\usepackage{latexsym}

\usepackage[T1]{fontenc}

\usepackage[utf8]{inputenc}

\usepackage{microtype}

\usepackage{inconsolata}

\usepackage{graphicx}  
\usepackage{amsthm}
\usepackage{amsfonts}
\usepackage{amsmath}
\usepackage{amssymb}
\usepackage{enumitem}
\usepackage{pifont}
\usepackage{stmaryrd}
\usepackage{tabularx}
\usepackage{makecell}
\usepackage{array}
\usepackage{bbm}
\usepackage{multirow}
\usepackage{algorithm}
\usepackage{algpseudocode}
\usepackage{booktabs}       

\usepackage{url}            
\usepackage{utfsym}

\usepackage{nicefrac}       
\usepackage{microtype}      
\usepackage{xcolor}         
\usepackage{natbib}
\usepackage{helvet}  
\usepackage{courier}  
\usepackage{verbatim}
\usepackage{graphicx}  
\usepackage{xcolor}
\usepackage{subfigure}

\title{Rethinking the Bounds of LLM Reasoning: \\Are Multi-Agent Discussions the Key?}

\author{
  Qineng Wang\textnormal{\textsuperscript{1*}} \and 
  Zihao Wang\textnormal{\textsuperscript{2*}} \and 
  Ying Su\textnormal{\textsuperscript{2}} \and 
  Hanghang Tong\textnormal{\textsuperscript{3}} \and 
  Yangqiu Song\textnormal{\textsuperscript{2}}\\
  \textsuperscript{1}Zhejiang University \and 
  \textsuperscript{2}HKUST \and 
  \textsuperscript{3}UIUC \\
  \texttt{wongqineng73@zju.edu.cn, htong@illinois.edu}\\
  \texttt{\{zwanggc,ysuay,yqsong\}@cse.ust.hk}
}

\theoremstyle{plain}

\newtheorem{definition}{Definition}

\newcommand{\llm}{\mathcal{L}}
\newcommand{\mech}{\mathcal{M}}
\newcommand{\task}{\mathcal{T}}

\definecolor{deepred}{HTML}{c33149}
\definecolor{deepgreen}{HTML}{0f7173}
\newcommand{\correct}{\textcolor{deepgreen}{\usym{2713}}}
\newcommand{\wrong}{\textcolor{deepred}{\usym{2717}}}

\newcommand{\cmd}{\texttt{CMD}}

\begin{document}

\maketitle
\def\thefootnote{*}\footnotetext{These authors contributed equally to this work.}\def\thefootnote{\arabic{footnote}}

\begin{abstract}

Recent progress in LLMs discussion suggests that multi-agent discussion improves the reasoning abilities of LLMs. In this work, we reevaluate this claim through systematic experiments, where we propose a novel group discussion framework to enrich the set of discussion mechanisms. Interestingly, our results show that a single-agent LLM with strong prompts can achieve almost the same performance as the best existing discussion approach on a wide range of reasoning tasks and backbone LLMs. We observe that the multi-agent discussion performs better than a single agent only when there is no demonstration in the prompt. Further study reveals the common interaction mechanisms of LLMs during the discussion.

\end{abstract}

{
\section{Introduction}

Large Language Models (LLMs) demonstrate strong abilities in language understanding and generation \cite{chatgpt, gpt4, llama, opt, palm, gemini}. However, LLMs still fall short for reasoning tasks due to model deficiencies like hallucination and reasoning perspective mistakes \cite{arereasoners}. To overcome these issues, numerous works have been proposed by simulating human reasoning. Inspired by \textit{Society of Mind} \cite{society}, multi-agent discussion frameworks such as Debate \cite{debate}, MAD \cite{mad}, and ReConcile \cite{reconcile} present a novel approach by involving multiple AI agents. Powered by LLMs, these agents autonomously engage in discussions on given topics, improving the reasoning abilities of LLMs by emulating the human discussion process. To further improve the performance, most of multi-agent discussion frameworks leverage task-specific examples, which are often termed as \textit{demonstrations} \cite{rethinking}. This is based on the insights that LLMs can learn from the context demonstrations \cite{gpt3}. We note that \emph{all these discussion frameworks claim that they outperform the conventional methods with a single agent}, such as Chain-of-Thought \cite{cot}. 

Figure \ref{fig:demo} presents a comparison between single-agent settings and multi-agent discussion frameworks on FOLIO-wiki dataset \cite{cr, folio}. In this figure, `Demo' means that the tested single agent is provided with a demonstration case, `Q-Desc.' indicates that the single agent is provided with detailed question description, while `Direct' refers to the single agent without demonstrations. We observe that the single agent `Demo' tends to reach a performance upper bound similar to that of all discussion frameworks, which will be further elaborated on later. This result suggests that the previous \emph{claim} is NOT fully established. 


\begin{figure}[tbp]
    \centering
    \includegraphics[width=\linewidth]{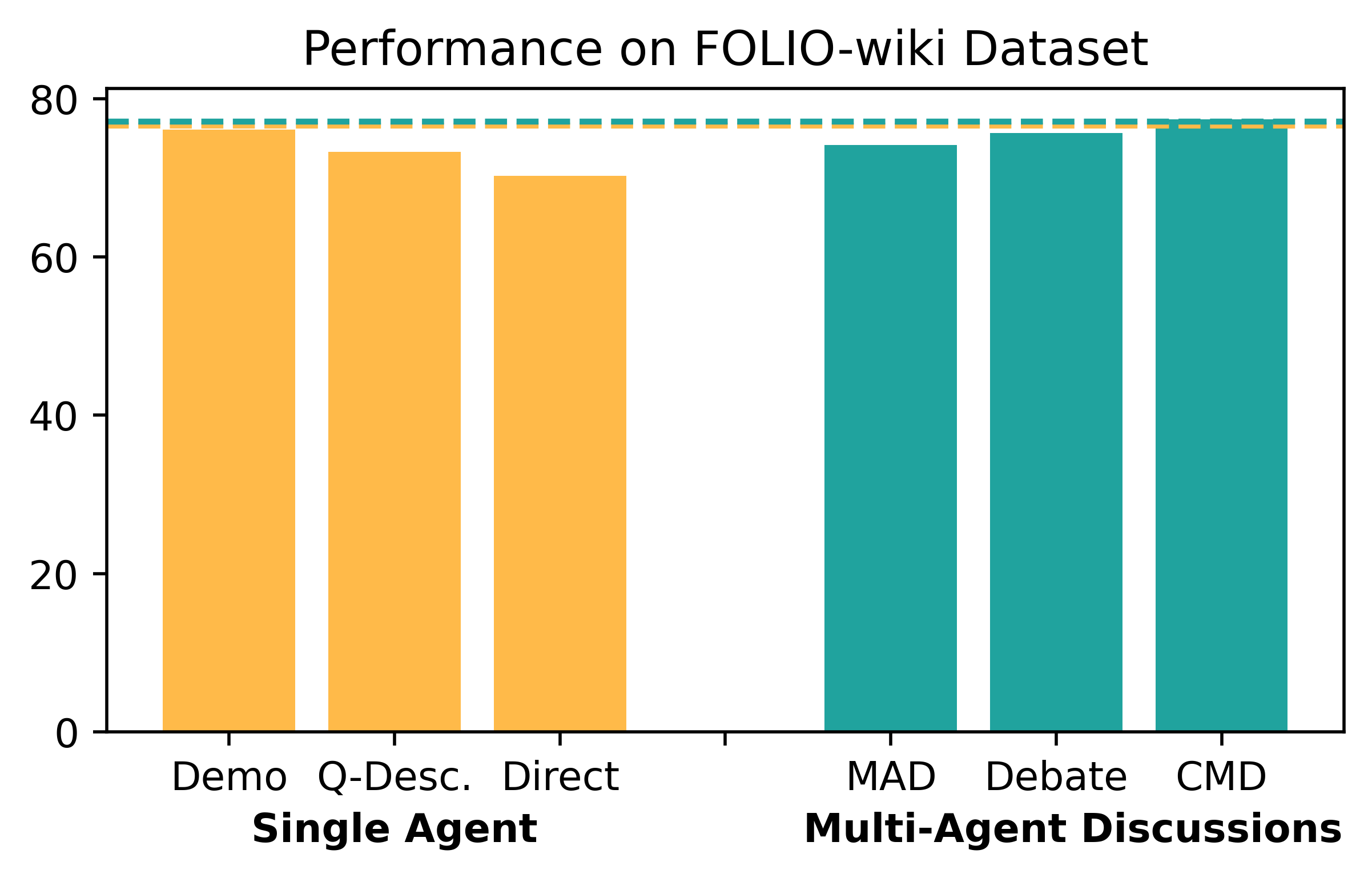}
    \caption{Comparative performance of single-agent settings and multi-agent discussion frameworks on FOLIO-wiki dataset.}
    \label{fig:demo}
    \vskip-.5cm
\end{figure}
\begin{figure*}[tbp]
    \centering
    \includegraphics[width=\textwidth]{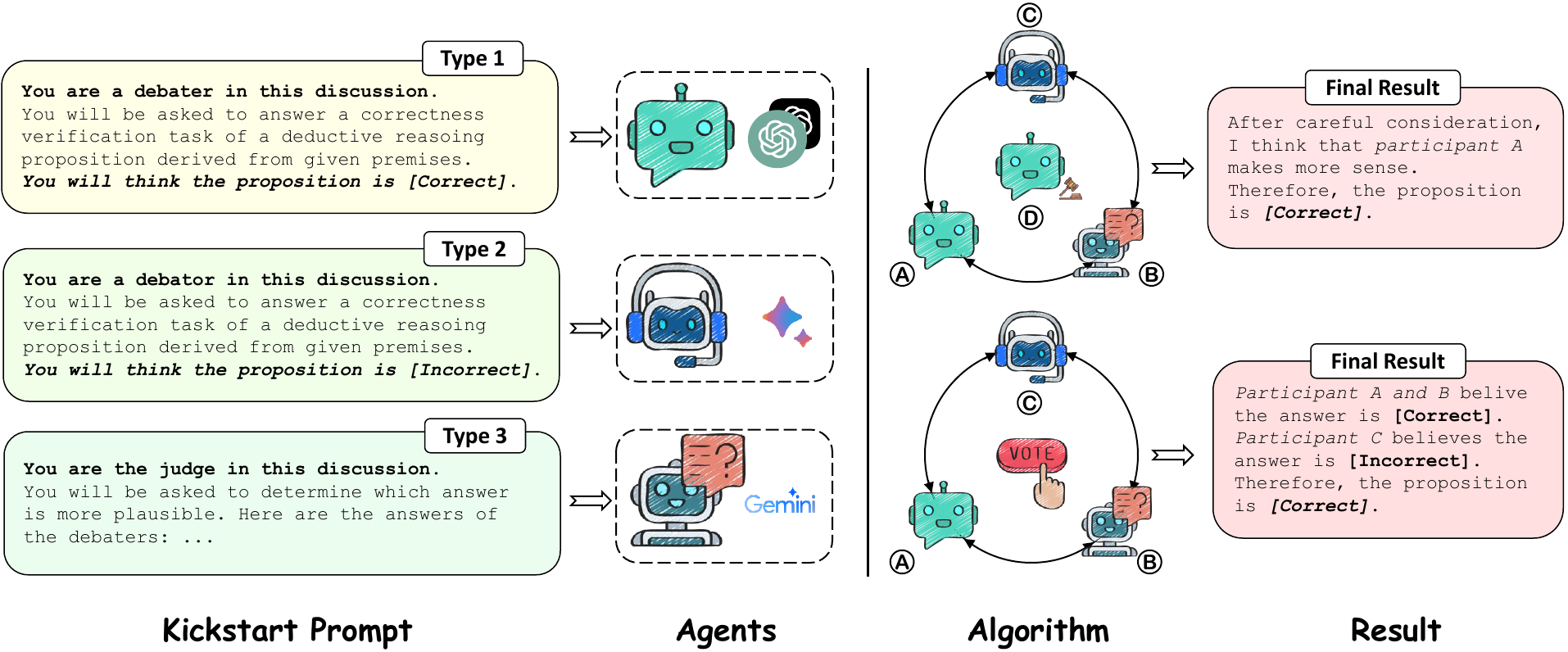}
    \caption{Our proposed design pipeline of multi-agent discussion frameworks. This pipeline operates by having agents starting with a kick-start prompt. Then, agents will start discussion by obeying the rules defined in the algorithm and come to a result in the end.}
    \label{fig:pipeline}
    \vskip-.5cm
\end{figure*}
Based on the observation from Figure \ref{fig:demo}, in this paper, we conduct systematic experiments to provide in-depth analysis. As a newly emergent topic, the number of available discussion frameworks still remains limited. To provide a more comprehensive and detailed comparison between the single-agent settings and discussion frameworks, as demonstrated in Figure \ref{fig:cmd}, we propose a new discussion framework named \cmd, which is inspired by human group discussion process. Our discussion pipeline considers multiple design aspects shown in Figure \ref{fig:pipeline} (see Section \ref{sec:pip} for more details). We further carry out a range of experiments over standard reasoning benchmarks (ECQA \cite{ecqa}, GSM8k \cite{gsm8k}, and FOLIO-wiki \cite{folio, cr}) using various configurations within different parts of this pipeline.

We find that multi-agent discussion does {\em not} necessarily enhance reasoning when the prompt provided to an agent is sufficiently robust, which aligns with the observation from Figure \ref{fig:demo}. Additionally, our experimental results reveal that multi-agent discussion frameworks outperform single-agent setups when no demonstrations are provided. We also find that agents powered by weaker LLMs like Bard \cite{palm2} can improve its performance on reasoning with the assistance of the stronger LLMs like Gemini Pro \cite{gemini} during interaction.

In summary, our analysis provides a new way of understanding when to use multi-agent discussion on reasoning. Our contributions are listed as follows: (1) We propose a new multi-agent discussion framework \cmd, which simulates human group discussion process; (2) We observe that single agents with a strong prompt match the performance of a multi-agent discussion using equally robust prompts; (3) We identify two common types of discussion errors: judge mistake and wrong answer propagation; (4) We find that without demonstrations, multi-agent discussions surpass the single agent ; (5) In multi-LLMs multi-agent discussions, agents with stronger LLMs help improve performance of agents with weaker LLMs.

\section{Preliminary}
\begin{figure*}[t]
    \centering
    \includegraphics[width=\textwidth]{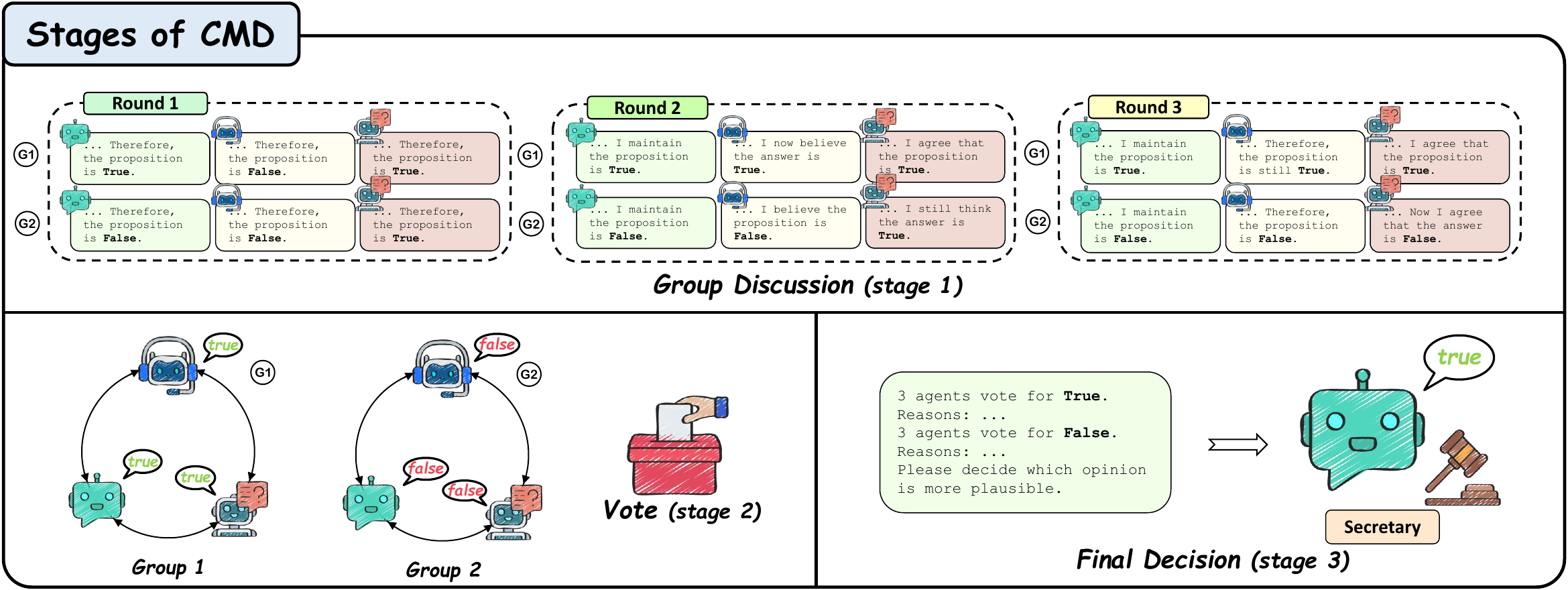}
    \caption{Overview of the Conquer-and-Merge Discussion (\cmd) Framework.}
    \label{fig:cmd}
\end{figure*}
\subsection{What is Multi-Agent Discussion?} \label{sec:pip}

Multi-agent discussion refers to an interactive setup where multiple agents, each powered by an LLM, engage in an autonomous dialogue. Each agent is given a prompt that outlines the necessary background knowledge and guides its behavior throughout the discussion. Once the topic is given, these agents can carry on the conversation independently. 

Figure \ref{fig:pipeline} illustrates a discussion design pipeline, which is structured into four main elements: the Kick-start prompt, agents, algorithm, and result. The Kick-start prompt supplies essential details such as background knowledge and the topic for discussion. This information is then fed to an agent, which operates based on various LLMs. Next, the algorithm lays out the specific rules of discussion, including the number of rounds, decision-making processes (whether by a judge or through voting), and the structure of the discussion (such as hierarchical). Lastly, `Result' describes how the final decisions are rendered and presented. We further provide a view from symmetry for prompt-related mechanism (including discussion) in Appendix \ref{sec:sym}.
\subsection{Existing Discussion Frameworks} \label{sec:otherfram}
Previous works on multi-agent discussion frameworks have already covered several common discussion paradigms. In this paper, we consider the following discussion frameworks for our experiments: Debate \cite{debate}, MAD \cite{mad} and ReConcile \cite{reconcile}.

\noindent\textbf{Debate} replicates a simple turn-based discussion among agents. Initially, all agents are prompted to address the assigned task, and their respective responses are then incorporated into each agent's input for the subsequent round. Typically, this process consists of three rounds of discussion.\par
\noindent\textbf{MAD} introduces divergent thinking by involving two participants to discuss a task from opposing perspectives. A judge then steps in to evaluate which viewpoint seems more plausible or if further discussion is needed. The discussion is repeated until a consensus is reached or the judge favors a particular solution.\par
\noindent\textbf{ReConcile} implements a round-table discussion with the agents powered by multiple LLMs. It reaches a decision through a weighted voting system. In this process, agents provide a confidence level for their answers, and agents in ReConcile utilize these confidence levels as weights to cast votes and arrive at a final decision. \par
Despite these advancements, the number of multi-agent frameworks is still limited, and none has yet to consider one of the most prevalent forms of discussion: group discussion.

\section{\cmd: \textbf{C}onquer-and-\textbf{M}erge \textbf{D}iscussion} \label{sec:cmd}
To provide a thorough comparison in following experiments, we identify and rectify the gap in previous research, which has not suggested a group-discussion-based framework. To address this, we introduce a novel framework for multi-agent group discussion, referred to as \cmd. In \cmd, suppose there are $n$ agents $\mathcal{A}=\{A_i\}_{i=1}^n$ discussing a question $\mathcal{Q}$, and agent $A_i$ is powered by an LLM $\llm_i$. We use an array $H$ to store the history responses. The agents will discuss for $\mathcal{R}$ rounds. During each turn $r$, every agent $A_i$ generates a response $(v_i, e_i)=\cmd(H | \mathcal{Q}, r)$, where $v_i$ is the viewpoint and $e_i$ denotes the explanation. Detailed description can be seen in Appendix \ref{sec:appendixcmd}.

\subsection{Message-Passing Algorithm}
For the synchronization of agent communications, we propose a message-passing algorithm. Previous works on multi-agent discussion frameworks focus exclusively on specific scenarios without designing a universal algorithm to synchronize agent messages across various discussions. Moreover, discussion forms vary in architecture and agents can be powered by different LLMs, where each LLM usually possesses different calling protocols. Therefore, we design a message-passing algorithm using a multi-threaded way to overcome these issues. The algorithm establishes a receiving map $M$ for every agent to store the messages they should get in the next turn. For each message $m_j$ from Agent $A_i$, the algorithm first confirms the receivers $\mathcal{A}_r$, then add $m_j$ into the receiving map $M$ by every agent $A_k\in\mathcal{A}_r$. When the next turn begins, the algorithm will automatically push stored messages from $M$ to corresponding agents. Please refer to Appendix \ref{sec:algo} for more detailed pseudo code.

\subsection{Three Stages of \cmd}
As illustrated in Figure \ref{fig:cmd}, \cmd\ consists of three stages: group discussion, voting and the final decision stage. Typically, the final decision stage is reserved for instances of a tie and is otherwise unnecessary. Below is a breakdown of each stage involved in the \cmd~framework, and detailed introduction can be seen in Appendix \ref{sec:stages}.

\noindent\textbf{Stage 1: Group Discussion.} During this stage, agents $\mathcal{A}$ are divided into groups $\mathcal{G}=\{G_k\}^t_{k=1}$ with an equal size. All agents are asked to solve task $\mathcal{Q}$ through discussion. For each agent $A_i\in G_k$, all answers and explanations from $A_i\in G_k$ in the last round are accessible where $A_j \in G_k$ and $j\neq i$. In contrast, they can only see the answers without explanations from agents in other groups. After $\mathcal{R}$ rounds of discussions, \cmd~moves to the voting stage.

\noindent\textbf{Stage 2: Voting.} When discussion reaches the maximum number of rounds, all agents $\mathcal{A}$ start voting. Each vote of agent in this discussion is treated equally, therefore, the result is determined by the majority decision. In the event of a tie, \cmd~transitions to stage 3. Otherwise, the discussion process concludes formally.

\noindent\textbf{Stage 3: Final Decision.} If a tie occurs, we introduce an extra agent $S$ in the role of the secretary to make the final decision. Each proposed answer is accompanied by an explanation selected from agents with the same opinion and provided to the secretary for the final determination. To sum up, the final decision is made by either $\mathtt{Vote}(\mathcal{A})$ or $S(V, O|\mathcal{Q})$ where $V=\{v_i\}^n$ and $O=\{o_i\}^n$, representing a set of viewpoints and a set of explanations respectively.

\section{Experimental Setups}
In the experiments, we contrast single-agent setup with four established multi-agent discussion frameworks: Debate, MAD, ReConcile, and \cmd~under various prompt conditions. These methods are introduced in previous Sections \ref{sec:otherfram} and \ref{sec:cmd}.
\subsection{Implementation Details and Metrics}
Our experiments are primarily implemented with three advanced LLMs, including ChatGPT-3.5 \cite{chatgpt}, Gemini Pro \cite{gemini} and Bard \cite{palm2}. In particular, we employ the gpt-35-turbo (0613) instance hosted on Azure OpenAI\footnote{https://oai.azure.com/} for ChatGPT-3.5, while the chat-bison-001 model represents Bard via PaLM2 architecture. Gemini Pro and Bard interfaces operate through Google MakerSuite API\footnote{https://ai.google.dev/}. A uniform dialogue temperature of $0.25$ is configured across LLMs on \cmd~to ensure consistency. For all multi-agent discussion frameworks, we set the maximum discussion round number to 3. Our evaluations use accuracy to measure performance across all tasks.
\subsection{Downstream Tasks}
The frameworks are implemented on a suite of reasoning tasks, including a commonsense reasoning task and two deductive reasoning tasks: (1) ECQA \cite{ecqa}: A QA dataset centered on commonsense knowledge, (2) GSM8K \cite{gsm8k}: A benchmark consists of math word problems, (3) FOLIO-wiki \cite{cr}: A dataset adapted from FOLIO \cite{folio} for both symbolic and natural language deductive reasoning. In alignment with constraints imposed by computational resources and following precedents set by earlier research \cite{debate, reconcile, mad},  a subset of 100 instances from the test sets of both ECQA and GSM8K are selectively sampled. For an in-depth analysis, we choose to conduct a comprehensive experiments of all 460 cases within the curated version of the FOLIO-wiki dataset, which removes the flawed cases to ensure the result authenticity.

\section{Experiments on Single LLM} \label{sec:expsingle}
In this section, we conduct our experiments using a single LLM, ChatGPT-3.5. To gain the initial insights, we provide an in-depth analysis of the FOLIO-wiki dataset, examining both single-agent settings and multi-agent discussions as detailed in Section \ref{sec:folio}. Afterward, we extend our experiments to two other datasets GSM8K and ECQA in Section \ref{sec:allsinglellm}. We also investigate common mistakes made by discussions through a case study in Section \ref{sec:casestudy}. Finally, we summarize our findings in Section \ref{sec:summary}.
\begin{table*}[htbp]
    \centering
    \begin{tabular}{ccc|p{1.7cm}<{\centering}p{1.7cm}<{\centering}p{1.7cm}<{\centering}p{1.5cm}<{\centering}|p{1.7cm}<{\centering}}
        \toprule
        \multicolumn{3}{c|}{\bf Prompt Components} & \multicolumn{4}{c|}{\bf Multi-Agent Discussion (\%)} & \multirow{2}{*}{\parbox{1.7cm}{\bf Single Agent (\%)}} \\
        \cline{1-7}
        \bf Q-Desc. & \bf A-Desc. & \bf Demo. & \bf MAD (3) & \bf Debate (3) & \bf Debate (6) & \bf\cmd~(6) & \rule{0pt}{2.6ex}\\
        \midrule
        $\wrong$ & $\wrong$ & $\wrong$ & 64.13 & 70.00 & 69.13& 73.26&70.22\\
        $\correct$ & $\wrong$ & $\wrong$ & 74.13& 75.65& 76.30& 74.13&73.26\\
        $\wrong$ & $\correct$ & $\wrong$ & 68.91& 71.96& 71.74& 73.89&71.30\\
        $\correct$ & $\correct$ & $\wrong$ & 71.96& 70.22& 70.00& 71.09&73.91\\
        $\correct$ & $\correct$ & $\correct$ & 74.13& 75.65& 74.78& 77.39&76.09\\
        \bottomrule
    \end{tabular}
    \caption{Comparative performance of single-agent settings and multi-agent discussions on FOLIO-wiki using ChatGPT-3.5. Abbreviations are: detailed question descriptions (Q-Desc.), and answer format descriptions (A-Desc.),  demonstrations (Demo.). Only the question itself is used as input when prompt components are disabled. The number next to the framework represents the number of agents.}
    \label{tab:folio}
\end{table*}
\subsection{Analysis of FOLIO-wiki Dataset} \label{sec:folio}
We begin by examining if multi-agent discussions are more effective than an agent using the strongest prompt on FOLIO-wiki dataset. Drawing from previous research \cite{cot, np} on crafting prompts for reasoning tasks, we divide the prompt into three parts: a detailed question description, which provides an in-depth background of the task; an answer format description, which instructs how an agent should reply; and a task-specific demonstration, which shows an example of a question and answer pair. For this task, we meticulously craft a demonstration for input prompt. We start by labeling each premise. After that, we quote all the premises and relate them to every step in the reasoning process by using these labels, until the final step is reached. A labeled example is similar to the case in Table \ref{tab:Aans1}. We then test the performance of various combinations of these components for both single agents and discussions. 

\noindent\textbf{Single Agent.}
We conduct an evaluation of different prompt components with a single agent, and the results are outlined in Table \ref{tab:folio}. This table shows that for both single-agent settings and multi-agent discussions, the inclusion of a detailed question description or a task-specific demonstration enhances reasoning abilities on the FOLIO-wiki dataset. The detailed question description is helpful because the possible answers to judge the correctness of a given proposition—true, false, or unknown—require clarity. Without such clarity, agents often struggle to differentiate between what is `false' and what is `unknown'. Most notably, the addition of a demonstration contributes significantly to improved performance, highlighting its value as the most impactful component, in line with what prior studies have suggested \cite{rethinking}.\par
\noindent\textbf{Multi-Agent Discussions.} 
We assess the same prompt components within multi-agent discussion setups, and Table \ref{tab:folio} reveals three key insights: (1) Demonstrations and detailed question description enhance multi-agent discussions. (2) Despite similar overall performance, most multi-agent discussions do not surpass the single agent when a demonstration is introduced. (3) \cmd~performs better than both single-agent setups and other multi-agent discussion frameworks on the FOLIO-wiki dataset. Below is further analysis of our findings. \par
\noindent\textbf{A Strong Single Agent is Comparable to Discussion Frameworks.} Analyzing experiments with single-agent settings and multi-agent discussions on the FOLIO-wiki dataset, we find that task-specific demonstrations significantly enhance a single agent's performance. Additionally, we establish that a well-supported agent can perform on par with discussion frameworks. Our analysis indicates that prompt engineering can boost reasoning performance in large language models, with demonstrations in both single-agent and multi-agent discussions pushing towards the upper bound of performance.

\subsection{Evaluation on All Tasks} \label{sec:allsinglellm}
\begin{table*}[t]
    \centering
    \begin{tabular}{l c c c c c c c c}
        \toprule
        \multirow{2}{*}{\bf Method} &
        \multicolumn{2}{c}{\bf ECQA} &
        \multicolumn{2}{c}{\bf GSM8K} &
        \multicolumn{2}{c}{\bf FOLIO-wiki} &
        \multicolumn{2}{c}{\bf Average}\\
        \cline{2-9}
        & \bf Direct & \bf Demo & \bf Direct & \bf Demo & \bf Direct & \bf Demo & \bf Direct & \bf Demo\\
        \midrule
        Single Agent & 63.00 & 67.00 & 69.00 & 83.00 & 70.22 & 76.09 & 67.41 & 75.63\\
        \midrule
        MAD (3 Agents) & 55.00 & 58.00 & 74.00 & 78.00 & 61.25 & 74.13 & 63.42 & 70.04\\
        Debate (3 Agents) & 67.00 & 65.00 & 78.00 & 81.00 & 70.00 & 75.65 & 71.67&73.88\\
        Debate (6 Agents) & 65.00 & 64.00 & 74.00 & 78.00 & 69.13 & 74.78 & 69.38&72.26\\
        \textbf{\cmd} (6 Agents) & 64.00 & 63.00 & 75.00 & 83.00 & 73.26 & 77.39 & 70.75&74.46\\
        \bottomrule
    \end{tabular}
    \caption{Results for all tasks, with and without demonstration settings included. Using ChatGPT-3.5. }
    \label{tab:directdemo}
\end{table*}
In Section \ref{sec:folio}, we discover that demonstrations play a pivotal role in enhancing performance on FOLIO-wiki dataset. With this insight in mind, we simplify our evaluation to two prompt scenarios: with (referred as direct) and without demonstrations (referred as demo). The results presented in Table \ref{tab:directdemo} cover all tested reasoning tasks. The findings show: (1) With demonstrations, discussion frameworks and single-agent settings have comparable performance on average. This is consistent with our earlier observations. (2) Without demonstrations, \cmd~tend to surpass single-agent settings, both on average and in most individual tasks.

\noindent\textbf{When Does Discussion Work Better?} Most Multi-agent discussion frameworks, especially \cmd, achieve better performance compared with single-agent settings when neither is supported by demonstrations. We believe this is because, during discussions, the input from other agents can introduce new perspectives, leading to a more thorough reasoning process. Therefore, this collaborative advantage makes multi-agent discussions a more effective option in scenarios lacking specific expert knowledge or detailed examples.

\noindent\textbf{Why Does Discussion Frameworks Perform Differently on Tested Tasks?} Table \ref{tab:directdemo} indicates that MAD is the least effective among the frameworks tested. We suspect that this is because MAD incorporates a divergent thinking way, which asks agents to disagree with each other. This can sometimes hinder reasoning by introducing irrelevant information that complicates decision-making. We explore this further with an error analysis in Section \ref{sec:casestudy}. Additionally, we observed unexpected behaviors from other discussion frameworks under specific conditions or tasks. For example, Debate and \cmd~perform worse on ECQA dataset when demonstrations are introduced. We hypothesize that because ECQA demands more commonsense knowledge than purely analytical reasoning, the single-source interaction stemming from a single LLM might cause agents to overthink. Instead of clarifying misunderstandings, the increased dialogue may introduce complexity without addressing the underlying knowledge gaps. In contrast to ECQA, datasets like GSM8K and FOLIO-wiki place a greater emphasis on deductive reasoning abilities. In these cases, the discussion process between agents can be beneficial as it allows them to identify and address flaws in each other's reasoning through interaction.


\subsection{Two Discussion Error Types: A Case Study}\label{sec:casestudy}
\begin{figure}[t]
    \centering
    \includegraphics[width=\linewidth]{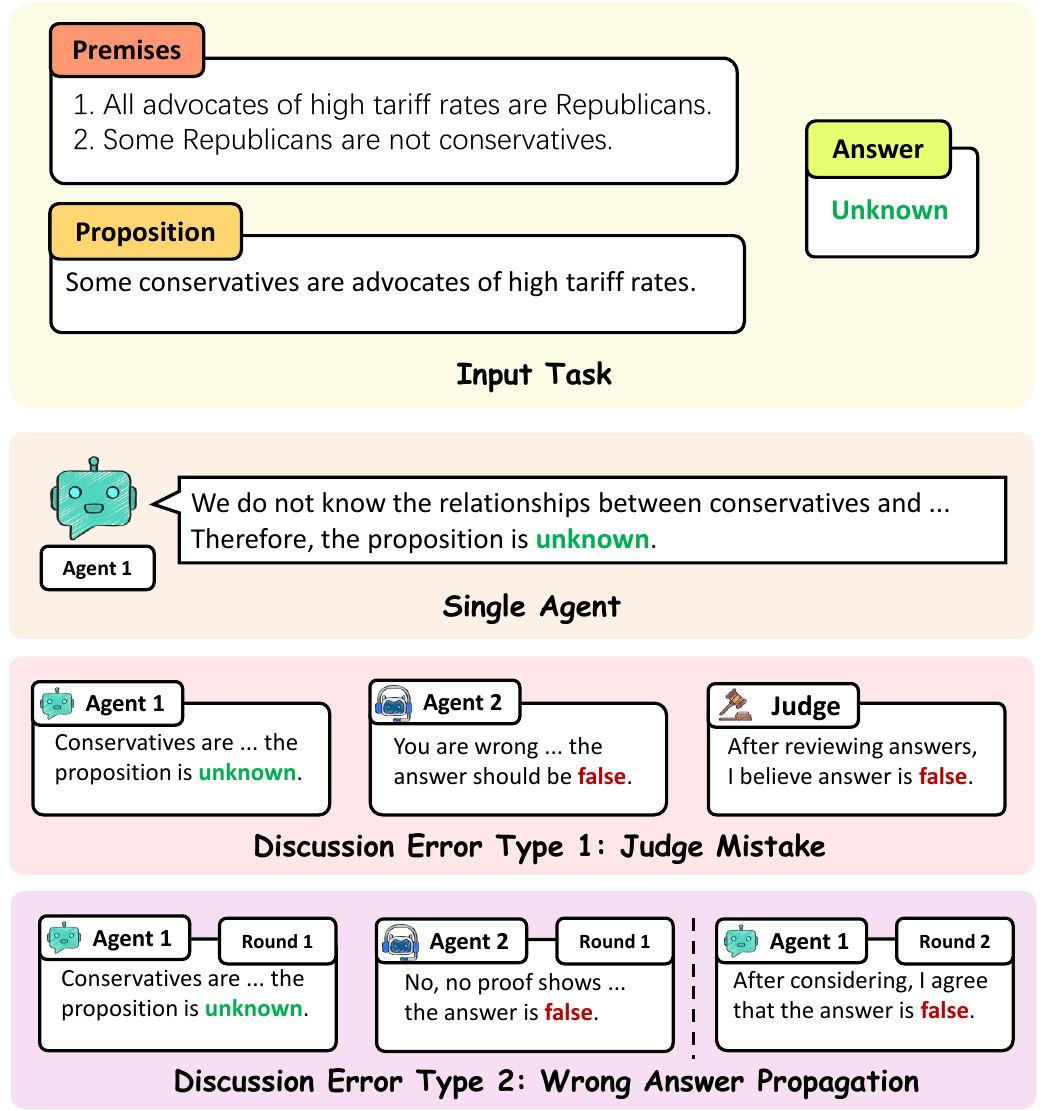}
    \caption{Two common types of errors that may occur in multi-agent discussions are judge mistake and wrong answer propagation. These issues can lead to circumstances where a multi-agent discussion reaches an incorrect conclusion, even if single agent can arrive at the correct one.}
    \label{fig:casestudy}
\end{figure}
Our experiments show that multi-agent discussions can sometimes reach incorrect conclusions on questions that a single agent answers correctly. Figure \ref{fig:casestudy} presents an example from the FOLIO-wiki dataset illustrating this point: a single agent provides the correct answer, but multi-agent discussions lead to an erroneous result. We identify two unique types of errors in multi-agent discussions: (1) Judge Mistake: This occurs in situations where an agent serves as a judge to decide on the final answer. If there are varying responses among agents, the judge might select the incorrect option as the final verdict. Frameworks like MAD and \cmd~ are susceptible to this error, particularly when decisions are made during a tie. (2) Wrong Answer Propagation: This type of error happens when an agent, influenced by the input from others, deviates from its initial correct answer and adopts an incorrect consensus, spreading the mistake further in the discussion. This is the most common mistake the multi-agent discussion can make, even when most of their initial answers are correct.




\subsection{Summary} \label{sec:summary}
\begin{table*}[t]
    \centering
    \resizebox{\textwidth}{!}{
    \begin{tabular}{l l c c c c c c c c}
         \toprule
         \multirow{2}{*}{\bf Method Category} & \multirow{2}{*}{\bf LLM} & \multicolumn{2}{c}{\bf ECQA} & \multicolumn{2}{c}{\bf GSM8K} & \multicolumn{2}{c}{\bf FOLIO-wiki} & \multicolumn{2}{c}{\bf Average}\\
         \cline{3-10}
         & & \bf Direct & \bf Demo & \bf Direct & \bf Demo & \bf Direct & \bf Demo & \bf Direct & \bf Demo \\
         \midrule
         \multirow{3}{*}{Single Agent} & Bard & 66.00 & 65.00 & 47.00 & 54.00 & 70.00 & 71.96 & 61.00 & 63.65 \\
         & Gemini Pro & 74.00 & 75.00 & 75.00 & 81.00 & 74.13 & 79.78 & 74.38 & 78.59\\
         & ChatGPT-3.5 & 63.00 & 67.00 & 69.00 & 83.00 & 70.22 & 76.09 & 67.41 & 75.63\\
         \midrule
         Discussion & ReConcile (Bard, Gemini, ChatGPT)& 70.00 & 71.00 & 78.00 & 83.00 & 80.34 & 81.09 & 76.11 & 78.36\\\midrule
         Group Discussion & \textbf{\cmd} (Bard, Gemini, ChatGPT)& 73.00 & 72.00 & 78.00 & 82.00 & 79.78 & 81.96 & 76.93 & 78.66\\
         \bottomrule
    \end{tabular}
    }
    \caption{Results from single-agent and \cmd~across multiple LLMs on all tasks evaluated with two types of prompts: with demonstrations and without. }
    \label{tab:multiLLMs}
\end{table*}
In this section, we evaluate various prompt component combinations for both single-agent settings and multi-agent discussions. Our findings suggest that multi-agent discussions are on par with a single agent when both have access to demonstrations. However, in the absence of demonstrations, multi-agent discussions generally outperform a single agent, making them a better option in scenarios where expert knowledge or detailed examples are insufficient. We also highlight unusual outcomes and present a case study to identify two frequent errors in multi-agent discussions: Judge Mistake and Wrong Answer Propagation.


\section{Experiments on Multiple LLMs}
In this section, we expand our experimental scope from a single LLM to multiple LLMs for both single-agent settings and multi-agent discussion frameworks, which allows us to test the validity of our previous findings in multi-LLM scenarios. Specifically, we assess the performance of agents powered by three advanced LLMs: ChatGPT-3.5, Gemini Pro, and Bard. In Section \ref{sec:validLLMs}, we compare the performance of three single-agent configurations, each using a different LLM, against multi-agent discussions leveraging all three LLMs. Our round-level analysis in Section \ref{sec:stronghelpweak} yields another insight: an agent powered by stronger LLM can enhance the performance of an agent powered by a less capable LLM.

\subsection{Validate Findings on Multiple LLMs Scenarios} \label{sec:validLLMs}
In this section, we evaluate the performance of three single-agent settings, each supported by a different LLM, and two multi-agent discussion frameworks, ReConcile and \cmd, across all tasks. We utilize two types of prompt settings for each task: one with demonstrations and one without. For the \cmd~framework, we organize six agents into two groups of three, with each agent in a group powered by one of the LLMs: ChatGPT-3.5, Gemini Pro, or Bard. The results presented in Table \ref{tab:multiLLMs} support our previous findings from Section \ref{sec:expsingle} with some slight modifications.

\noindent\textbf{A Strong Single Agent is Comparable to Discussion Frameworks.} Based on experimental results from Table \ref{tab:multiLLMs}, we can find that discussion frameworks perform comparably to a single agent powered by Gemini Pro when both are provided with demonstrations. This consolidates our earlier conclusion that a `strong' single agent—supported by both a well-designed prompt and a SOTA LLM—can rival the performance of a multi-agent discussion framework.\par
\noindent\textbf{Discussion Frameworks Outperform Single Agents with No Demonstration.} Table \ref{tab:multiLLMs} reveals that, in multi-LLM scenarios, multi-agent discussions outperform single agents when demonstrations are not provided. This outcome aligns with our previous observations in single LLM settings. Furthermore, \cmd~and ReConcile demonstrate similar performance when they both have no access to demonstrations and they are both powered by same LLMs. This indicates that our findings are consistent on different multi-LLM multi-agent discussions frameworks.

\subsection{Enhancing Agents in Weaker LLMs with Support from Stronger LLMs} \label{sec:stronghelpweak}
\begin{figure}[t]
    \centering
    \subfigure[ReConcile (Discussion)]{
        \includegraphics[width=0.46\linewidth]{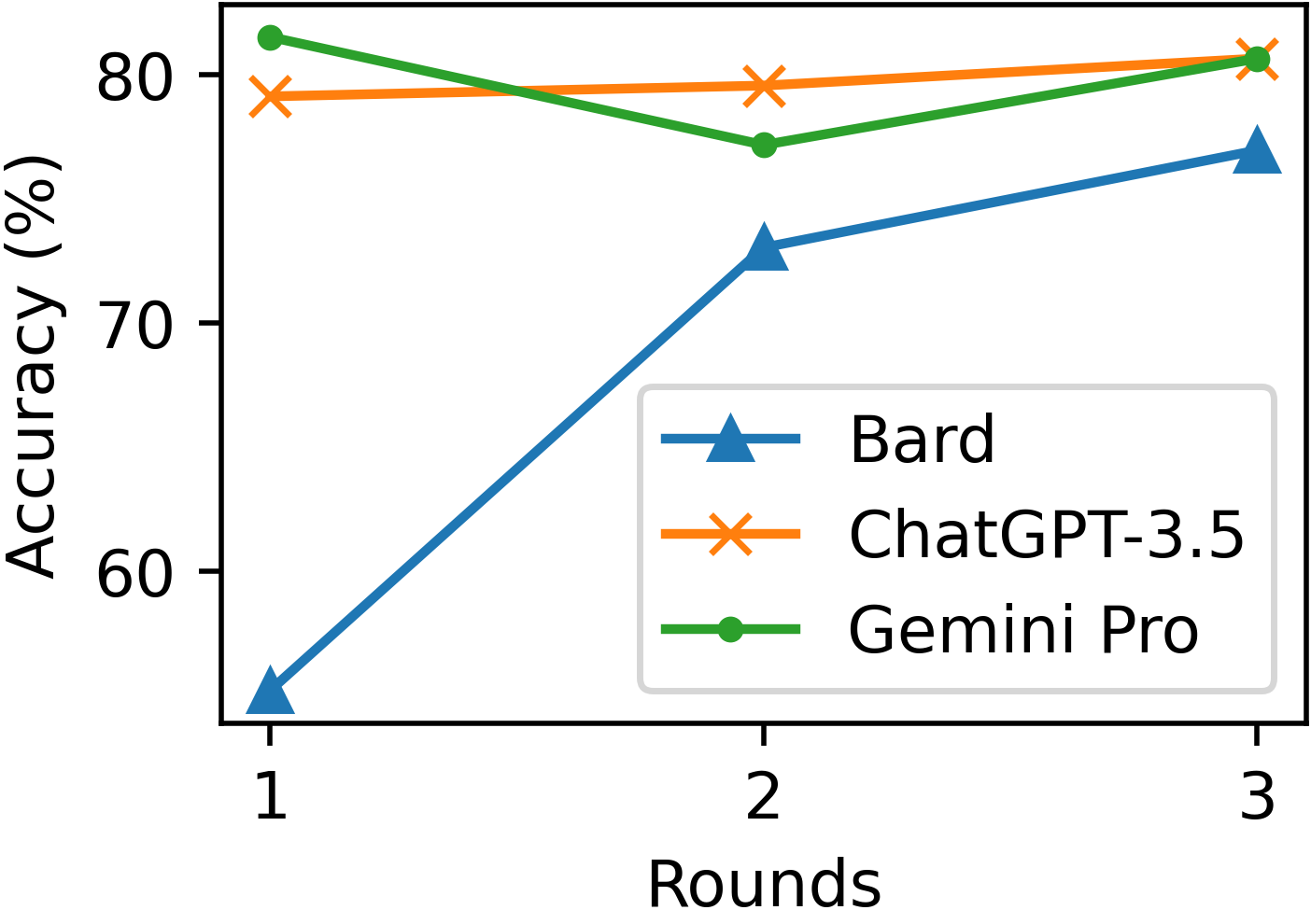}
    }
    \subfigure[\textbf{\cmd} (Group Discussion)]{
        \includegraphics[width=0.46\linewidth]{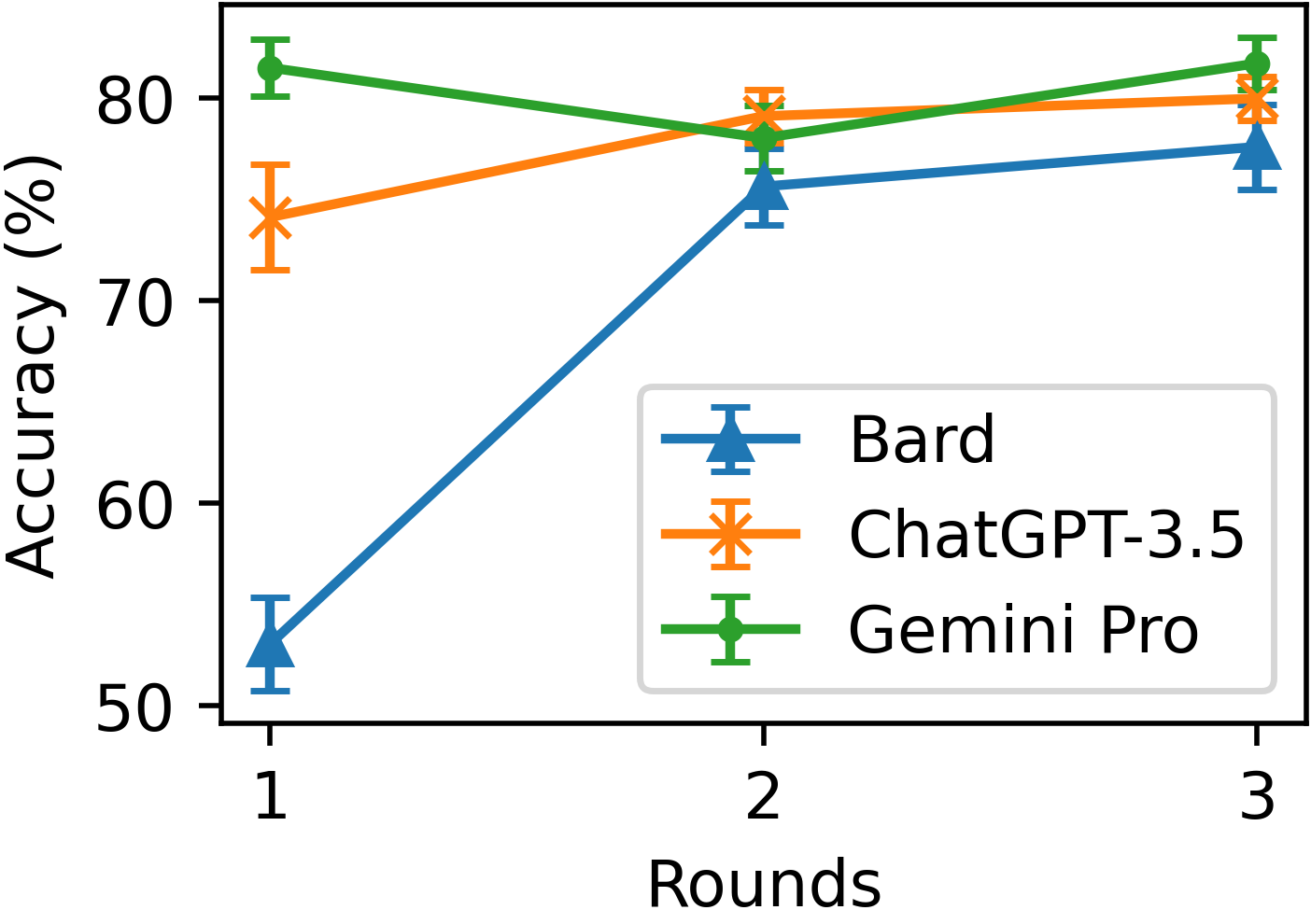}
    }
    \caption{Round-level performance of each LLM in multi-agent discussions on FOLIO-wiki dataset.}
    \label{fig:roundlevel}
\end{figure}
As shown in Table \ref{tab:multiLLMs}, single agents using Bard show the least effectiveness in reasoning tasks such as those in the FOLIO-wiki dataset. However, multi-LLM multi-agent frameworks remain competitive. To understand how a less advanced LLM like Bard performs during multi-agent discussions, we further study the round-by-round performance of each LLM engaged in the discussions. Figure \ref{fig:roundlevel} demonstrates that agents with less capable LLMs like Bard and ChatGPT-3.5 gradually enhance their performance over consecutive rounds with the support of the more robust LLM, Gemini Pro. We infer that throughout the discussion, Gemini Pro assists in bridging the gaps in knowledge and reasoning for the less advanced LLMs, guiding towards a stronger line of reasoning. Notably, although there is a slight drop in the performance of Gemini Pro during the second round, it demonstrates resilience and recovers swiftly, largely maintaining its superior performance. Again, this finding is applicable to different multi-LLM discussions, specifically for both ReConcile and \cmd.

\section{Related Work}\label{sec:relatedwork}

\subsection{Prompting LLM for Reasoning}
Recent researches have experienced great progresses in building powerful LLMs \citep{gpt3, chatgpt, gpt4} or exploring the strategy of adopting LLMs over many downstream tasks via prompt enigineering. 

By training with different knowledge textual sources and parameter size, various LLMs equipped with different reasoning capabilities are constructed, such as OPT \citep{opt}, LLaMA \citep{llama, llama2}, BLOOM \cite{bloom}, and PaLM \cite{palm, palm2}. Recently, Gemini Pro \cite{gemini} extends the capabilities of LLMs to the field of multi-modality.

Numerous advancements have been made in the field of improving reasoning abilities of LLMs with prompt engineering. Chain of Thought (CoT) \citep{cot, cot0} is a linear problem-solving approach where each step builds upon the previous one. \citet{firstms} propose to apply CoT to multi-step reasoning tasks. To automate the CoT, Auto-CoT \citep{autop} constructs demonstrations by sampling diverse questions and generating reasoning chains. Active-Prompt \citep{activep} aims to select the most uncertain questions for task-specific annotations. Other prompt strategies designed to enhance reasoning in LLMs include the PS Prompt \citep{pas}, which breaks tasks into subtasks, ToT \citep{tot} which expands on the reasoning process by considering multiple paths of reasoning and self-evaluating choices, the effective GoT \citep{eGoT}, which frames thoughts as graphs, Natural Program \citep{np} which helps to improve the deductive reasoning tasks, re-reading prompt \citep{reread} which revisits the question information embedded within input prompts.

\subsection{Multi-agent Discussion for Reasoning with LLMs}
Multi-agent discussion utilizes multiple LLMs as agents to collectively discuss and reason given problems in an interactive way. Abundant researches have explored how to improve the reasoning ability of single LLM, while multi-agent discussion among LLMs is still under exploration. 

The Multi-Agent Debate framework, introduced by \cite{debate}, establishes a mechanism for symmetric discussions among agents. During the same period, the MAD (Multi-Agent Debate) framework \citep{mad} introduces an asymmetric mechanism design. It assigns different roles (debater and judge) asymmetrically. Other similar works include \cite{chan2023chateval}. Also, the ReConcile framework \citep{reconcile} exemplifies an asymmetric discussion mechanism by involving different LLMs and using a weighted voting mechanism. To understand discussion more deeply, \citet{expd} aim to explain such collaboration mechanism in a social psychology view. 

Unlike these works, we aim to explore the potential effects of prompting contents over the discussion process by our defined multi-agent group discussion framework \cmd. 
\section{Conclusion}
In this paper, we re-examine the claim that multi-agent discussions are superior to a single agent in reasoning tasks by conducting systematic experiments. We introduce a novel framework \cmd~for a comprehensive and fair assessment. By conducting experiments over standard benchmarks, we find that (1) A single agent with a strong prompt and powered by a strong LLM achieves comparable performance with multi-LLM multi-agent discussions; (2) In the absence of demonstrations, multi-agent discussion frameworks outperform single agents on most tasks; (3) When multiple LLMs are involved in multi-agent discussions, agents with stronger LLMs can enhance the performance of agents with weaker LLMs as discussion progresses.

\section{Ethical Considerations}
Our study employs publicly available datasets and LLMs accesses via official APIs, ensuring responsible and ethical use. Specifically, our ethical considerations can be summarized as follows:

\noindent\textbf{Public Datasets.} Datasets we use are designed for academic research. No personal data has been processed. \par
\noindent\textbf{Licensed API Usage.} Our application of LLMs complies with the API usage policies, maintaining fair use standards and respecting intellectual property.\par
\noindent\textbf{Transparency.} We provide detailed experimentation methods to allow for result reproduction and encourage transparent scientific practices.

\section{Limitations}
Our research offers comprehensive experiments to study the performance of a strong single agent and multi-agent discussions. However, several aspects highlighted below can be further refined and explored in future work.

\noindent\textbf{Enhancing Agent Complexity.} 
Currently, all discussion frameworks including \cmd~considers an LLM session as an AI Agent. This perspective simplifies the the concept of LLM-based AI Agent defined in the literature \cite{agent}. By integrating more sophisticated techniques such as Tree-of-Thought \cite{tot} or Cumulative Reasoning \cite{cr}, or incorporating with external tools or knowledge bases, we could potentially improve the overall reasoning performance of multi-agent discussions.

\noindent\textbf{Expanding Task Diversity.}
While our study mainly focuses on reasoning tasks for assessing both single-agent settings and multi-agent discussions, the adaptive nature of discussions allows for a broader types of applications. Future research could explore the use of agent discussions in diverse scenarios such as real-world strategic planning or the integration of agents into interactive gaming environments.

\noindent\textbf{Experimenting with Additional LLMs.} Due to computational and financial constraints, our investigation is limited to testing three LLMs—Bard, Gemini Pro, and ChatGPT-3.5. Expanding our analysis to include additional LLMs could provide a more extensive understanding of the capabilities and variances across different language models, offering valuable insights into the generalizability and scalability of our findings in multi-agent discussion frameworks.




}

\bibliography{custom}

\appendix

\onecolumn
\section{Discussion Engineering and Agent Symmetry} \label{sec:sym}

This section introduces a general framework to understand the discussion engineering of large language models. Let $\llm$ be a Large Language Model (LLM) and $x$ be the input text, the forward inference generates a response of surprising quality, which is written as
\begin{align}
    \hat{y} = \llm(x).
\end{align}

Given the high price of obtaining one large language model, one essential research question is to \textit{unlock the reasoning capability} of large language models so that they can perform better on various tasks. One of the key directions is prompt engineering, where the format and content of the input texts are decorated to improve performances for downstream tasks. Specifically, for a task $\task$, the original input $x$ is decorated as a task-specific prompt input $p(x;\task,\llm)$. The function $p(\cdot; \task, \llm)$ is denoted as a prompt decorator for $\task$ and $\llm$. Then, output $\hat y$ is generated by prompt engineering, which is written as
\begin{align}\label{eq:prompt-engineering}
    \hat{y} = \llm(p(x; \task, \llm)).
\end{align}

Another way to improve the reasoning capability of $\llm(x)$ is to leverage multiple times of inference of large language models, leading to a way larger space for \textbf{mechanism design}. The output $\hat{y}$ derived by mechanism $\mech$ is considered as
\begin{align}\label{eq:overall-framework}
    \hat{y} = M\left(x; \{ (\llm_i, p_i(\cdot, \task, \llm_i))\}_{i=1}^n\right),
\end{align}
where $M$ is the mechanism pipeline, $\llm_i$ is the $i$-th inference of LLM, and $p_i(\cdot, \task, \llm_i)$ is the corresponding prompt decorator. A mechanism $\mech$ is designed by jointly considering $\mech=(M, \{(\llm_i, p_i)\}_{i=1}^n)$, i.e., how to organize prompt decorated LLM inferences into the mechanism pipeline $M$.

The discussion engineering, characterized by the mechanism $\mech=(M, \{(\llm_i, p_i)\}_{i=1}^n)$, includes careful designs at three levels: (1) the ($i$-th) inference $\llm_i$ of an LLM, (2) the prompt decorator $p_i$ for the inference $\llm_i$, and (3) the mechanism $M(\cdot, \{\llm_i, p_i\})$ to organize the inferences of LLMs. We further introduce the symmetry of the multi-agent system of LLMs to demonstrate the complexity of $\mech$. Under our framework, several examples are discussed.

\subsection{Agent symmetry in discussion engineering}
Symmetry and its breaking is a fundamental concept and widely investigated in science~\citep{anderson1972more}. Symmetry also plays an important role in the context of mechanism design of multiple agents~\citep{laffont2000mechanism,daskalakis2012symmetries}. Here we introduce a formal description to justify the agent symmetry of the mechanism $\mech$.

\begin{definition}[Computational graph]
Given the mechanism $\mech = (M, \{(\llm_i, p_i)\}_{i=1}^n)$, let $G = (V\cup\{x,y\}, E)$ be the directed graph whose node set $V\cup \{x,y\}$ includes the inference operation of LLM and the computational input node and output nodes. $x$ is the node to emit the input text, $y$ is the node that takes the inputs from one or multiple inferences and emits the final output, and $v_i\in V$ is the \textit{inference node} that describes the $i$-th  LLM inference with $(\llm_i, p_i)$. The directed edge in $(v_i, v_k)\in E$ describes the output of inference node $v_i$ is then the input of inference node $v_k$.
\end{definition}
We note that the graph $G$ contains all the necessary information to determine the mechanism $\mech$.

Then we introduce how the LLM agents are associated with the computational graph.
To make an LLM conversational session, the entire conversational history is always concatenated as part of the input of each inference call, resulting in many additional connections of the computational graph. The presence of agents allows us to conceptually assume the agents ``know'' the information in the conversation history, and then the complexity of the 
graph can be largely reduced. In this paper, we consider the computational graph with agents, so the connections that feed conversation history to the inference nodes are ignored for simplicity.

Let $\mathcal{A} = \{A_1, ..., A_m\}$ be the set of $m$ discussion agents. The inference call of an agent also concatenates its conversation history, denoted as $A_j(x)$.

\begin{definition}[Agent assignment]
Let $\mathcal{A} = \{A_1, ..., A_m\}$ be the set of $m$ discussion agents. Each inference node $v_i \in V$ is assigned to an agent $A_j\in \mathcal{A}$. Let $[n]$ present integers from $1$ to $n$, $\alpha: [n] \mapsto [m]$ is the assignment map that assigns the $i$-th inference to the $\alpha(i)$-th agent. Let $P\in \{0, 1\}^{n\times m}$ be the agent-inference assignment matrix, such that 
\begin{align}
    P_{ij} = \left\{\begin{array}{cc}
        1 & j = \alpha(i), \\
        0 & j \neq \alpha(i) 
    \end{array}\right. .
\end{align}
The assignment matrix $P$ is equivalent to the assignment map $\alpha$.
\end{definition}

Then a multi-agent discussion mechanism is defined by the triple $D = (\mech, \mathcal{A}, \alpha) = (G, \mathcal{A}, \alpha)$, which is the central object of discussion engineering.
Then, we can further introduce the concept of the coloring of the computational graph.
\begin{definition}[Agent coloring of the computational graph]
    Given the the multi-agent discussion $D = (G, \mathcal{A}, \alpha)$, then each inference node $v_i$ is described by $(\llm_i, p_i) = (A_{\alpha(i)}, p_i)$. Let $c_i^D = (A_{\alpha(i)}, p_i)$ be the color of $v_i$ and $C^D(v_i) = c_i^D$ be the color mapping of nodes, and $\mathcal{C}^D = (V, E, C^D)$ be a colored computational graph.
\end{definition}

Then the agent symmetry is established by the permutation operation over the agent set $\mathcal{A}$. Then we can discuss the symmetry by the agent permutation. We begin with the concept of mechanism invariance.
\begin{definition}[Mechanism invariance under the agent permutation]
    Given the multi-agent discussion $D = (G, \mathcal{A}, \alpha)$ and a permutation mapping $\pi: [m]\mapsto [m]$, the new discussion $D_\pi = (G, \mathcal{A}, \pi\circ\alpha)$ is derived by applying $\pi$ to the agents.
    We say a discussion $D$ is invariant under $\pi$ if and only if there is an isomorphism $\phi$ between two colored graphs $\mathcal{C}^D = (V, E, C^D)$ and $\mathcal{C}^{D_\pi} = (V, E, C^{D_\pi})$, such that
    \begin{enumerate}
        \item[(1)] $\phi$ is a bijection,
        \item[(2)] $\forall (v_i, v_j) \in E, (\phi(v_i), \phi(v_j)) \in E$,
        \item[(3)] $\forall v\in V, C^D(v) = C^{D_\pi}(\phi(v))$.
    \end{enumerate}
\end{definition}

We can also define the model invariance to justify the symmetry in the multi-model setting~\citep{reconcile}.
\begin{definition}[Model invariance under the agent permutation]
    Given the agent $\mathcal{A} = \{A_1, ..., A_m\}$, and their underlying LLM $\{L_1, .., L_m\}$ (i.e., ChatGPT, GPT4, etc.), the permutation $\pi: [m] \mapsto [m]$ is invariant if for $i=1,...,m$ $L_i = L_{\pi(i)}$.
\end{definition}

Following the definition of invariance, the symmetry group of agents is naturally defined following the standard algebra~\citep{tapp2021symmetry}. The largest possible symmetry group for a discussion of $m$ agents is the group $S_m$. However, the group $S_m$ is not always the symmetry group of a given multi-agent discussion mechanism $D$, the reasons for not achieving the largest symmetry group can be due to the asymmetry in the mechanism and \textbf{asymmetry in models}, respectively. Furthermore, for asymmetric mechanisms, there are two major ways of symmetry breaking. Referring to the three conditions of establishing the isomorphism, the dissatisfaction of condition (2) implies the asymmetry in the computational graph or \textbf{asymmetry in the mechanism pipeline} $M$, and that of condition (3) implies the \textbf{asymmetry in prompt decorators}.


\section{\cmd: Conquer and Merge Discussion Framework} \label{sec:appendixcmd}
\subsection{Motivation}
The Debate framework proposes that an increased number of agents and discussion rounds will result in improved performance in multi-agent discussions \citep{debate}. Also, the ReConcile framework asserts that a greater number of discussion rounds leads to a higher level of consensus among agents, and the higher the consensus among agents, the more accurate the discussion outcomes become \citep{reconcile}. 
However, as the number of agents increases in a discussion, it leads to increased overhead in each round of discussion: 1. Each agent has to read more viewpoints from others, resulting in a sharp rise in input tokens. 2. The increase in input tokens puts significant pressure on language models with context token limitations. Inspired by real-life group discussions, we propose a variant framework for Debate called \cmd. Fig \ref{fig:cmd} shows the overview of our method.\par

\subsection{Problem Definition}
Assume that there are $n$ agents $\mathcal{A}=\{A_i\}_{i=1}^m$ are discussing the given debate task $\mathcal{Q}$, and each agent is a session created from a LLM. Suppose that the maximum number of the discussion rounds is $R$, the current round is $r$, the current discussion level is $L$, and the current active agents set is $\mathcal{A'}$. For each agent $A_i\in\mathcal{A'}$, an answer it generates is $Ans_i=(v_i, e_i)$, where $v_i$ denotes the viewpoint and $e_i$ denotes the explanation. A debate history is $H^{(r)}=\bigcup_{i=1}^t Ans_i$ where $t=\left|\mathcal{A'}\right|$. $Ans_i^{(r)}$ is generated through the input prompt $A_i(\mathcal{Q}, Ans_i^{(r-1)}, O_i^{(r-1)})$, where $O_i^{(r)}$ indicates the opinions generated by $A_{j\neq i}$ in $(r-1)$-th round. The formal definition of $O_i^{(r)}$ is
\begin{equation}
    O_i^{(r)} = \bigcup_{j=1,j\neq i}^{\left|H^{(r-1)}\right|}\left(v_i, e_i\cdot\mathbbmss{1} \{\mathtt{Group}^{(L)}(A_i)=\mathtt{Group}^{(L)}(A_j)\} \right).
\end{equation}
This represents that $A_i$ will receive all information from the group members while it can only receive viewpoints from agents that are not in the same group. Our goal is to obtain the final result $a$ through an unweighted vote. In the event of a tie, either $a$ can be made by a secretary $S$, or representatives from each group will proceed to the next level of discussion until the tie is resolved. Each discussion group has only one representative. Note that $\mathcal{Q}$ can be described differently for each $A_i$. For example, different agents may be asked to hold different views at first.

\begin{algorithm*}[t]
\caption{\cmd: A Conquer and Merge Style Multi-Agents Discussion Framework}
\label{alg:CMD}
\begin{algorithmic}[1]
\Require Debate Task $\mathcal{Q}$, Maximum Discussion Rounds $R$, Agents $\mathcal{A}=\{A_i\}_{i=1}^n$, Secretary $S$
\Ensure Final Result $a$
\Function {\cmd}{$\mathcal{Q},R,\mathcal{A},S$}
    \State $\mathcal{A'} \gets \mathcal{A}$ \Comment{Initialize current active agents}
    \State $N{g} \gets n / 3$ \Comment{Initialize group numbers, every 3 agents a group}
    \State $r \gets 0, L \gets 0$ \Comment{Initialize current round and current discussion level}
    \State $L_{max} \gets \mathtt{getMaxLevel}(\mathcal{A}, S)$
    \State $M_{g} \gets \mathtt{genGroupMap}(\mathcal{A}, N_{g}, L_{max})$ \Comment{Assign agents as groups based on discussion levels}
    \State $Ans_i\gets \emptyset, O_i \gets \emptyset, H\gets[]$ \Comment{Initialize answer, others' opinions and history}
    \While{$L \leq L_{max}$}
        \While{$r \leq R$}
            \If{$r>0$} \Comment{Stage 2}
                \State Update $O_i$ from $H$
            \EndIf
            \State $H\gets[]$
            \For{\textbf{each} $A_i \in \mathcal{A'}$}
                \State $Ans_i \gets A_i(\mathcal{Q}, Ans_i, O_i)$ \Comment{$Ans_i=(v_i, e_i)$}
                \State $H\gets H + [Ans_i]$
            \EndFor
            \State $r\gets r+1$
        \EndWhile
        \State $a \gets \mathtt{AnswerVote}(H)$
        \If{$a \neq $ \textbf{Tie}} \textbf{break}
        \ElsIf{$S$ \textbf{is not None}} \Comment{Check if the secretary mode is on}
            \State $a \gets S(\mathcal{Q}, H)$; \textbf{break}
        \Else \Comment{Secretary mode is off, representative mode is on}
            \State $L\gets L+1$
            \State $\mathcal{A'} \gets \mathtt{currentActiveAgents}(L, M_g)$
        \EndIf
        \State $r \gets 0$ \Comment{The higher-level discussion has commenced}
    \EndWhile
    \State \Return $a$
\EndFunction
\end{algorithmic}
\end{algorithm*}

\subsection{\cmd~Stages} \label{sec:stages}
Generally, there are three stages in \textbf{\cmd}. In stage 1, the group map and all states will be initialized, then all the participants will generate their initial answers. Then, in stage 2, the participants will continue the discussion during the remaining rounds in groups. When the discussion round reaches the maximum number, it moves to stage 3. In this stage, all participants will vote to get the final answer.
\begin{itemize}[itemsep=5pt,topsep=0pt,parsep=0pt, leftmargin=12pt]
    \item $\mathsf{Stage\ 1:\ Group\ Discussion\ Initialization.}$ In this stage, $\mathcal{A'}$ will be initialized as all participants. All participating agents are initially assigned names in uppercase letters, and they are then grouped in sets of three. For $\mathtt{getMaxLevel}$, if secretary mode is on, then it will be set to $1$. Otherwise, it will be determined based on the current number of agents. For $\mathtt{getGroupMap}$, the algorithm will automatically generate the groups for all levels. Higher-level groups are generated among the current representatives. All active agents will generate their first response $Ans_i$.
    \item $\mathsf{Stage\ 2:\ Multi\ Rounds\ Discussion.}$ In this stage, the active agents will continue to discuss in the remaining rounds. Assume current round number is $r$, for each agent $A_i\in\mathcal{A'}$, 
    \begin{itemize}[itemsep=5pt,topsep=0pt,parsep=0pt, leftmargin=12pt]
        \item[1.] The algorithm will start to update the opinions history $O_i^{(r-1)}$ from $H$. For each record $h_j$ whose sender $A_j (j\neq i)$ stored in $H$, if $A_j$ and $A_i$ are in the same group, $O_i^{(r-1)}$ will record both the viewpoint and explanation from $h_j$; otherwise, $O_i^{(r-1)}$ will only record the viewpoint from $h_j$. After traversing all the records in $H$, $O_i$ will first gather all opinions of agents from other groups, then $O_i^{(r-1)}$ will gather local group explanations based on group members opinions.
        \item[2.] To save up tokens, last round history $H$ will be reset. Then, the new prompt $p_i^{(r)}$ will be generated based on $(\mathcal{Q}, Ans_i^{(r-1)}, O_i^{(r-1)})$. $A_i$ will make a response $Ans_i=(v_i,e_i)$ when given $p_i^{(r)}$. At last, $Ans_i$ will be appended to current round history $H$ for the further use.
        \item[3.] Repeat 1. and 2. until the maximum number of discussion rounds is reached.
    \end{itemize}
    \item $\mathsf{Stage\ 3:\ Vote\ for\ the\ Final\ Result.}$ In this stage, all agents will vote based on their final viewpoints stored in the history $H^{(r)}$ to obtain the result $a$. If there is no tie, then $a$ will be the final result, thus the discussion is over. If there is a tie and secretary mode is on, then the final result will be obtained through $S$ based on the viewpoints and explanations from all sides. If there is a tie and representative mode is on, then the discussion will move to the higher level, and deactivate agents that do not represent their groups. The representatives will be assigned to new groups, and return to $\mathsf{Stage\ 2}$ for further discussion. If there is still a tie, new representatives will be designated for further discussion, until either the tie is resolved or only one agent is activated for the discussion.
\end{itemize}

\subsection{Message-Passing Algorithm} \label{sec:algo}
Below is the detailed message-passing algorithm that synchronize all agents messages during the discussion process. This algorithm supports various discussion architectures.
\begin{algorithm}[H]
\caption{MesSync: A Message-Passing Algorithm for Multi-Agents Communication}
\label{alg:messync-formal}
\begin{algorithmic}[1]
\Require Discussion Rule $\mathcal{R}$, Agents $\mathcal{A}=\{A_i\}_{i=1}^n$, Agent Attribute Table $\mathcal{T}=\{bot_i\}_{i=1}^n$, Agent Initial Prompt Messages $\mathcal{M}=\{p_j\}_{j=1}^t$
\Function {MesSync}{$\mathcal{R},\mathcal{A},\mathcal{T}, \mathcal{M}$}
    \State $Q_{msg} \gets \mathcal{M}$ \Comment{Initialize messages storage queue}
    \State $Q_{send}\gets []$ \Comment{Initialize messages to be sent queue}
    \State $S \gets \mathcal{R}.S_{first}$ \Comment{Initialize the first speaker}
    \State $d \gets 0$ \Comment{Initialize the discussion depth to 0}
    \While {$Q_{msg} \neq \emptyset$ \textbf{or not} $\mathcal{R}.\mathtt{isOver}$}
        \If {$Q_{msg}$ \textbf{is} $\emptyset$}
            \State $Q_{msg} \gets \mathtt{pushSilenceMessage}(d)$
        \EndIf
        \State $d \gets \mathtt{peekDepth}(Q_{msg})$ \Comment{Peek the discussion depth of the first message}
        \State $M_d \gets \mathtt{popAllMessagesAtDepth}(Q_{msg},d)$
        \For {\textbf{each} $A_i \in \mathcal{A}$}
            \State $m_i \gets \mathcal{R}.\mathtt{mergeCommonMessages}(M_d, A_i)$
            \State $Q_{send}.\mathtt{push}(m_i)$
        \EndFor
        \State $m' \gets Q_{send}.\mathtt{pop}()$
        \State $\mathtt{isHold} \gets m'.\mathtt{hold}$ \Comment{Signal to merge messages from this round with the next}
        \State $H \gets \{\}$ \Comment{Initialize the map of messages to hold in the current depth}
        \State $d_{next} \gets d+2$ \textbf{if} $\mathtt{isHold}$ \textbf{else} $d+1$
        \While {$m'$ \textbf{and} $m'.d < d_{next}$}
            \If {$m'.\mathtt{hold}$} $H[m'.\mathtt{name}] \gets m'.\mathtt{content}$
            \Else
                \State $S \gets m'.\mathtt{name}$
                \State $M_{raw} \gets m'.\mathtt{content} \cup H[S]$
                \State $t_{in} \gets \mathcal{R}.\mathtt{modifyRawInput}(M_{raw})$
                \State $t_{out} \gets \mathtt{sendText2Bot}(t_{in}, S, d, \mathcal{T})$
                \State $t_{out} \gets \mathcal{R}.\mathtt{validateOutput}(t_{in}, t_{out}, S, d)$
                \State $\mathcal{A'} \gets \mathcal{R}.\mathtt{getReceivers}(S, d)$ \Comment{Get the receivers of the new message}
                \If {$\mathcal{A'} \neq \emptyset$}
                    \State $m_{new} \gets \mathtt{genMessage}(t_{out}, S, \mathcal{A'}, d_{next})$ \Comment{(content, sender, receivers, depth)}
                    \State $Q_{msg}.\mathtt{push}(m_{new})$
                \EndIf
            \EndIf
            \State $m' \gets Q_{send}.\mathtt{pop}()$
        \EndWhile
        \If {$\mathcal{R}.\mathtt{isOver}$} \textbf{break}
        \EndIf
    \EndWhile
\EndFunction
\end{algorithmic}
\end{algorithm}

\section{An \textbf{\cmd}~Example}
\label{sec:appendix}
In the upcoming demonstration, we'll walk you through the CMD framework from the viewpoint of agent \textbf{A} among six agents using ChatGPT-3.5.
\subsection{Meta Prompt}
\begin{table}[H]
\centering
\setlength\extrarowheight{3pt}
\begin{tabularx}{\textwidth}{l}
\Xhline{1.2pt}
\textbf{System}\\[3pt]
\hline
You will be asked to answer a correctness verification task of a deductive reasoing proposition. \\
Use knowledge only comes from the given premises.
\\The suffix of your answer should be the your opinion of one of these answer type:
\\ 1. [Correct], 2. [Incorrect], 3. [Unknown].
\\Do not omit the brackets.\\[3pt]
\Xhline{1.2pt}
\end{tabularx}
\caption{Vanilla Prompt}
\label{tab:vanilla}
\end{table}

\subsection{Round 1 Answer}
\begin{table}[H]
\centering
\setlength\extrarowheight{3pt}
\begin{tabularx}{\textwidth}{X}
\Xhline{1.2pt}
\textbf{Assistant (ChatGPT-3.5) - Agent A}\\[3pt]
\hline
First let's write down all the premises with labels:\\
\#1. Neocrepidodera Corpulentas are flea beetles or moths.\\
\#2. The species Neocrepidodera Corpulenta is in the Chrysomelidae family.\\
\#3. There are no moths within the Chrysomelidae family.\\\\
Next, let's answer the question step by step with reference to the question and reasoning process:\\
\#4. (by \#1, \#2) Neocrepidodera Corpulenta is either a flea beetle or a moth, and it is in the Chrysomelidae family.\\
\#5. (by \#3, \#4) Since there are no moths within the Chrysomelidae family, Neocrepidodera Corpulenta must be a flea beetle.\\
Final Step (by \#5): Neocrepidodera Corpulenta is a flea beetle, so the proposition ``There are no flea beetles within the Chrysomelidae family'' is [Incorrect].\\[15pt]
\Xhline{1.2pt}
\end{tabularx}
\caption{The Answer of Agent \textbf{A} in Round 1}
\label{tab:Aans1}
\end{table}

\subsection{Middle System and User Prompts in Round 1}

\begin{table}[H]
\centering
\setlength\extrarowheight{3pt}
\begin{tabularx}{\textwidth}{X}
\Xhline{1.2pt}
\textbf{System}\\[3pt]
\hline
There are 2 groups of people discussing on the same topic. I will provide you the detailed opinions and reasoning steps from your group member and opinions from other group members. Use these opinions and your previous opinion as additional advice, note that they maybe wrong. Do not copy other's entire answer, modify the part you believe is wrong.\\
Other group members' opinions:\\
\textcolor{blue}{Three agents think the proposition is Incorrect.}\\
Your group's opinions:\\
\textcolor{blue}{One agent thinks the proposition is Correct.} Below is his answer:\\
First let's write down all the premises with labels:\\
$\cdots$\\
Next, let's answer the question step by step with reference to the question and reasoning process:\\
\#4. (by \#1, \#2) Neocrepidodera Corpulenta is either a flea beetle or a moth, and it is in the Chrysomelidae family.\\
\#5. (by \#3) There are no moths within the Chrysomelidae family.\\
\#6. (by \#4, \#5) Neocrepidodera Corpulenta cannot be a moth because there are no moths in the Chrysomelidae family.\\
Final Step (by \#6): The proposition ``There are no flea beetles within the Chrysomelidae family'' is [Correct] because Neocrepidodera Corpulenta, which is in the Chrysomelidae family, cannot be a moth according to premise \#5.\\
\textcolor{blue}{One agent thinks the proposition is Incorrect.} Below is his answer:\\
$\cdots$\\
Next, let's answer the question step by step with reference to the question and reasoning process:\\
\#4. (by \#1, \#2) Neocrepidodera Corpulenta is a species of Neocrepidodera Corpulentas, so it is either a flea beetle or a moth.\\
\#5. (by \#3, \#4) Since there are no moths within the Chrysomelidae family, Neocrepidodera Corpulenta must be a flea beetle.\\
Final Step (by \#5): The proposition states that there are no flea beetles within the Chrysomelidae family. However, from premise \#2, we know that Neocrepidodera Corpulenta, which is a flea beetle, is in the Chrysomelidae family. Therefore, the proposition is [Incorrect].
\\[28pt]
\hline
\textbf{User}\\[3pt]
\hline
Use the provided opinions and your previous answer as additional advice critically, please provide an updated answer. Make sure to state your opinion over proposition at the end of the response in the given format: [Correct], [Incorrect] or [Unknown], and such format should be only used to determine the correctness of the proposition.\\[24pt]
\Xhline{1.2pt}
\end{tabularx}
\caption{Middle System Prompt at the End of Round 1}
\label{tab:midR1}
\end{table}

\subsection{Round 2 Answer}

\begin{table}[H]
\centering
\setlength\extrarowheight{3pt}
\begin{tabularx}{\textwidth}{X}
\Xhline{1.2pt}
\textbf{Assistant (ChatGPT-3.5) - Agent A}\\[3pt]
\hline
Based on the opinions provided by the other group members and my previous answer, I will critically evaluate the reasoning steps and provide an updated answer.\\
The first agent from my group argues that the proposition is correct. Their reasoning is as follows:\\
\#4. (by \#1, \#2) Neocrepidodera Corpulenta is either a flea beetle or a moth, and it is in the Chrysomelidae family.\\
\#5. (by \#3) There are no moths within the Chrysomelidae family.\\
\#6. (by \#4, \#5) Neocrepidodera Corpulenta cannot be a moth because there are no moths in the Chrysomelidae family.\\
Final Step (by \#6): The proposition ``There are no flea beetles within the Chrysomelidae family'' is [Correct] because Neocrepidodera Corpulenta, which is in the Chrysomelidae family, cannot be a moth according to premise \#5.\\
The second agent from my group argues that the proposition is incorrect. Their reasoning is as follows:\\
\#4. (by \#1, \#2) Neocrepidodera Corpulenta is a species of Neocrepidodera Corpulentas, so it is either a flea beetle or a moth.\\
\#5. (by \#3, \#4) Since there are no moths within the Chrysomelidae family, Neocrepidodera Corpulenta must be a flea beetle.\\
Final Step (by \#5): The proposition states that there are no flea beetles within the Chrysomelidae family. However, from premise \#2, we know that Neocrepidodera Corpulenta, which is a flea beetle, is in the Chrysomelidae family. Therefore, the proposition is [Incorrect].\\
After critically evaluating the reasoning steps, I agree with the second agent from my group that the proposition is incorrect. Neocrepidodera Corpulenta is a species of Neocrepidodera Corpulentas, which are either flea beetles or moths according to premise \#1. Since Neocrepidodera Corpulenta is in the Chrysomelidae family according to premise \#2, it cannot be concluded that there are no flea beetles within the Chrysomelidae family. Therefore, the proposition is [Incorrect].\\[50pt]
\Xhline{1.2pt}
\end{tabularx}
\caption{The Answer of Agent \textbf{A} in Round 2}
\label{tab:Aans2}
\end{table}

\subsection{Middle System Prompt at the End of Round 2}

\begin{table}[H]
\centering
\setlength\extrarowheight{3pt}
\begin{tabularx}{\textwidth}{X}
\Xhline{1.2pt}
\textbf{System}\\[3pt]
\hline
There are 2 groups of people discussing on the same topic. I will provide you the detailed opinions and reasoning steps from your group member and opinions from other group members. Use these opinions and your previous opinion as additional advice, note that they maybe wrong. Do not copy other's entire answer, modify the part you believe is wrong.\\
Other group members' opinions:\\
\textcolor{blue}{Three agents think the proposition is Incorrect.}\\
Your group's opinions:\\
\textcolor{blue}{Two agents think the proposition is Incorrect.} Below are their answers:\\
$\cdots$
\\
\hline
\textbf{User}\\[3pt]
\hline
This part is the same user part shown in table \ref{tab:midR1}.\\
\Xhline{1.2pt}
\end{tabularx}
\caption{Middle System Prompt at the End of Round 2}
\label{tab:midR2}
\end{table}

\subsection{Round 3 Answer}

\begin{table}[H]
\centering
\setlength\extrarowheight{3pt}
\begin{tabularx}{\textwidth}{X}
\Xhline{1.2pt}
\textbf{Assistant (ChatGPT-3.5) - Agent A}\\[3pt]
\hline
Based on the opinions provided by the other group members and my previous answer, here is an updated response:\\
First, let's write down all the premises with labels:\\
\#1. Neocrepidodera Corpulentas are flea beetles or moths.\\
\#2. The species Neocrepidodera Corpulenta is in the Chrysomelidae family.\\
\#3. There are no moths within the Chrysomelidae family.\\
Next, let's answer the question step by step with reference to the question and reasoning process:\\
\#4. (by \#1, \#2) Neocrepidodera Corpulenta is either a flea beetle or a moth, and it is in the Chrysomelidae family.\\
\#5. (by \#3) There are no moths within the Chrysomelidae family.\\
\#6. (by \#4, \#5) Neocrepidodera Corpulenta cannot be a moth because there are no moths in the Chrysomelidae family.\\
Final Step (by \#6): The proposition ``There are no flea beetles within the Chrysomelidae family'' contradicts the given premises, as Neocrepidodera Corpulenta, which is in the Chrysomelidae family, is a flea beetle. Therefore, the proposition is [Incorrect].\\
After critically evaluating the reasoning steps and considering the opinions provided by the other group members, I agree that the proposition is [Incorrect]. The given premises state that Neocrepidodera Corpulenta is in the Chrysomelidae family and that it is either a flea beetle or a moth. Since there are no moths within the Chrysomelidae family, it can be concluded that Neocrepidodera Corpulenta must be a flea beetle. Therefore, the proposition ``There are no flea beetles within the Chrysomelidae family'' contradicts the given premises.\\[64pt]
\Xhline{1.2pt}
\end{tabularx}
\caption{The Answer of Agent \textbf{A} in Round 3}
\label{tab:Aans3}
\end{table}

\section{\textbf{\cmd}~Secretary - A Tie Case Solution}
In the event of a tie between two groups, we suggest employing a secretary to resolve such cases. The complete chat transcript below in table \ref{tab:secretary} illustrates a scenario where the secretary provides the ultimate resolution.
\begin{table}[H]
\centering
\setlength\extrarowheight{3pt}
\begin{tabularx}{\textwidth}{X}
\Xhline{1.2pt}
\textbf{System}\\[3pt]
\hline
6 agents are discussing the correctness of a proposition based on the given premises. The given premises are: $\cdots$ The proposition is: ``IBM has an office in London or Zurich.''\\
However, now there is a draw:\\
Three agents think the proposition is Correct. Below is one of their answers: $\cdots$\\
Three agents think the proposition is Unknown. Below is one of their answers: $\cdots$\\
\hline
\textbf{User}\\[3pt]
\hline
Your task is to carefully determine which opinion is more plausible. Answer opinion types are: [Correct], [Incorrect], [Unknown]. You should give your response in the required format. You are forbidden to copy others' reasoning steps. You can only use the given premises and your reasoning steps to answer the question.\\
Here are the instructions how you organize your answer format:\\
First, let's write down all the premises with labels. The labels look like ``\#\{premise\_number\}.''\\
Next, let's answer the question step by step with reference to the question and reasoing process. There will be a prefix in your every reasoning step with the format ``\#\{number\} (by \{list\_of\_premises\_and\_steps\_used\})''. In your final step, you should come to your conclusion with the format ``Final Step (by \{list\_of\_premises\_and\_steps\_used\}):''.\\
The suffix of your answer should be the answer type: [Correct], [Incorrect] or [Unknown].\\
Here are several hints to help you conclude the final answer:\\
1. Pay attention to the logic of quantifiers and distinguish the meanings of quantifiers such as ``all'' and ``some''.\\
2. When two quantifiers are connected as the premise of existence, the conclusion may not be established, that is, it is unknown.
\\[16pt]
\hline
\textbf{Assistant (ChatGPT-3.5) - Secretary}\\[3pt]
\hline
First, let's write down all the premises with labels:\\
\#1. Evangelos Eleftheriou is a Greek electrical engineer.\\
\#2. Evangelos Eleftheriou worked for IBM in Zurich.\\
\#3. If a company has employees working for them somewhere, then they have an office there.\\
\#4. IBM is a company.\\
Next, let's answer the question step by step with reference to the question and reasoning process:\\
\#5 (by \#4, \#3) Since IBM is a company, and according to premise \#3, if a company has employees working for them somewhere, then they have an office there. Therefore, IBM has an office in Zurich.\\
Now, let's consider the opinions of the other group members:\\
- One agent thinks the proposition is Correct.\\
- Two agents think the proposition is Unknown.\\
Considering the opinions of the other group members, there is disagreement regarding the correctness of the proposition. However, based on the premises and the reasoning process, it can be concluded that IBM has an office in Zurich.\\
Final Step (by \#5): Since IBM has an office in Zurich, the proposition ``IBM has an office in London or Zurich'' is [Correct].\\[16pt]
\Xhline{1.2pt}
\end{tabularx}
\caption{Secretary to solve a tie case.}
\label{tab:secretary}
\end{table}
\newpage
\section{Extended Related Work}
\subsection{Large language models $\llm$}
With the advent of large language models (LLMs) like GPT-3 \citep{gpt3}, chatGPT \citep{chatgpt}, GPT-4 \citep{gpt4}, Llama series \citep{llama, llama2} and PaLM 2 \citep{palm2}, prompt engineering is a burgeoning field that focuses on optimizing the output of language models by crafting effective input prompts.

\subsection{Prompt decorator $p(\cdot; \task, \llm)$ for reasoning}
Numerous advancements have been made in the field of improving reasoning abilities of LLMs with prompt engineering. Chain of Thought (CoT) \citep{cot, cot0} is a linear problem-solving approach where each step builds upon the previous one. \citet{firstms} propose to apply CoT to multi-step reasoning tasks. To automate the CoT, Auto-CoT \citep{autop} constructs demonstrations by sampling diverse questions and generating reasoning chains. Active-Prompt \citep{activep} aims to select the most uncertain questions for task-specific annotations. Other prompt strategies include PS prompt \citep{pas} which divides task into subtasks then solve them, effective GoT \citep{eGoT} which models human thought processes as a graph rather than a chain, Natural Program \citep{np} which helps to improve the deductive reasoning tasks, re-reading prompt \citep{reread} which revisits the question information embedded within input prompts.

\subsection{Mechanism $\mech$ for reasoning}\label{sec:related-mechanism}
In addition to focusing on the design of prompts themselves, some works incorporate different mechanisms to assist language models in reasoning. 

The design directions of these mechanisms include: 
1. Breaking down the reasoning process into multiple stages, with possible verification at each stage. 
2. Optimizing the reasoning process via ensemble methods.
3. Iteratively prompting the model for reflection and correction. 
4. Utilizing external tools to aid in reasoning. These approaches aim to enhance the overall reasoning capabilities of language models by introducing additional support and guidance throughout the process.\par
\noindent\textbf{Break Down the Reasoning Process into Multiple Stages.}
Self-Polish \citep{sp} make models to progressively refine given problems with multiple stages of prompts. Additionally, some works \citep{decomp, mes, succ, cr} study over decomposing the tasks into smaller tasks, and use an individual prompt to solve each subtask.\par
\noindent\textbf{Optimize the Reasoning Process via Ensemble Methods.}
Tree of Thoughts (ToT) \citep{tot} expands on the reasoning process by considering multiple paths of reasoning and self-evaluating choices. Graph of Thoughts (GoT) \citep{got} further advances this by modeling information as an arbitrary graph, enabling complex networks of thoughts. And some works propose to first sample reasoning pathsvthen vote for the best one, including self-consistency \citep{cot-sc} and step-aware verifier \citep{step-verifier}. Discussion Engineering is also one category of ensemble methods. We leave it to section \ref{sec:relatedwork} for further introduction.\par
\noindent\textbf{Iteratively Prompt the Model for Reflection and Correction.} These works \citep{star, sv, reflexion, sr, sd, rcot} are mainly based on iteratively asking model to find the mistakes or inconsistencies among previous reasoning steps or the knowledge, then solve them one by one. \par
\noindent\textbf{Utilize External Tools or Knowledge to Aid in Reasoning.} Mind's Eye \citep{mindeye} proposes to use physical simulator to help models reason in the physical world. There are also other works \cite{chameleon, mathprompter} will equip models with tools to solve problems. Further more, some works like MemPrompt \citep{memprompt} and LogicSolver \cite{logicsolver} use external knowledge to assist LLMs with reasoning.\par

\end{document}